# ᛕ: A Universal Measure of Predictive Intelligence

David Gamez

*Department of Computer Science, Middlesex University, London, UK*
*d.gamez@mdx.ac.uk / www.davidgamez.eu*

**Abstract**

Over the last thirty years, considerable progress has been made with the development of systems that can drive cars, play games, predict protein folding and generate natural language. These systems are described as intelligent and there has been a great deal of talk about the rapid increase in artificial intelligence and its potential dangers. However, our theoretical understanding of intelligence and ability to measure it lag far behind our capacity for building systems that mimic intelligent human behaviour. There is no commonly agreed definition of the intelligence that AI systems are said to possess. No-one has developed a practical measure that would enable us to compare the intelligence of humans, animals and AIs on a single ratio scale.

This paper sets out a new universal measure of intelligence that is based on the hypothesis that prediction is the most important component of intelligence. As an agent interacts with its normal environment, the accuracy of its predictions is summed up and the complexity of its predictions and perceived environment is accounted for using Kolmogorov complexity. Two experiments were carried out to evaluate the practical feasibility of the algorithm. These demonstrated that it could measure the intelligence of an agent embodied in a virtual maze and an agent that makes predictions about time-series data. This universal measure could be the starting point for a new comparative science of intelligence that ranks humans, animals and AIs on a single ratio scale.

## 1. Introduction

In recent years there have been breakthroughs in the ability of computers to recognize faces, drive cars, play games, predict protein folding and engage in natural language conversation. These systems are typically described as artificially intelligent and doom-laden predictions have been made about the existential threat of AIs with superhuman levels of general intelligence. Behind all this hype is a poorly defined property called "intelligence", which has only been statistically measured in a few well-defined populations. To evaluate the progress we have made with AI and assess its potential dangers we need a clear definition of intelligence and a way of measuring it that will enable us to compare the intelligence of humans, animals[1] and artificial systems on a single ratio scale.

      The nature of intelligence has been extensively discussed, and many definitions have been put forward. Psychologists have linked intelligence to cognitive ability, rational thinking, problem-solving and understanding. In the AI community intelligence is often defined in terms of an agent's ability to achieve goals and receive rewards from its environment. This paper puts forward several reasons for thinking that intelligence is closely connected to an agent's ability to make accurate predictions. These include theories about the predictive brain, the use of prediction in planning, the prediction component in intelligence tests, and the role that prediction plays in AI systems.

---

[1] To improve readability, this paper will refer to non-human animals simply as "animals".

People often assume that humans have the same level of intelligence in all environments. A person's IQ is not said to be lower when they are eating or higher when they are delivering a lecture on quantum theory. However, several observations suggest that people's problem-solving skills vary with their environment. For example, human intelligence is particularly weak in high-dimensional environments and in environments that consist of large quantities of numerical data. If human intelligence is narrow, there is little reason to believe that we could create general intelligence in a machine. A more reasonable position is that all the intelligences we know (humans, animals and AIs) are, to a greater or lesser extent, specialized for specific environments. If this is correct, measures of intelligence should index an agent's level of intelligence to a set of environments in which it has this level of intelligence.

Discussions about intelligence also lack clarity about what is meant by an environment, which is often naively assumed to be the physical world. However, agents can only apply their intelligence to what they perceive, and different agents can experience the same physical environment in radically different ways. For example, suppose that a human, dog and Khepera robot are in a room. The dog cannot read a map on the wall, the Khepera only has two-dimensional distance information, and none of the agents has anything intelligent to say about neutrino patterns in the room. This issue can be ignored when comparing agents with similar senses, such as humans. It cannot be ignored when comparing systems with different senses and diverse ways of processing sensory data into higher level representations.

The interpretation of intelligence that is set out in this paper is the starting point for a new universal measure of intelligence, which is based on an agent's internal state transitions. As the agent explores its environment, the accuracy of its predictions is summed up for each perceived environment. Trivial predictions are eliminated using Kolmogorov complexity and intelligence across multiple environments is combined using a second Kolmogorov complexity component. Finally, the logarithm is taken to make it easier to compare simple and complex systems.

Two experiments were carried out to test the practical feasibility of this algorithm. The first measured the intelligence of an agent embodied in a maze environment. As the agent moves through a maze it learns to make statistical predictions about the consequences of its actions at different locations. The second experiment measured the intelligence of an agent that uses a neural network to predict future values of time-series data. These experiments demonstrated that the algorithm is a practical way of measuring predictive intelligence in medium-sized AI systems.

The first part of the paper gives some background about the measurement of intelligence in humans, animals and AIs and describes previous work on universal measures of intelligence. The next section advances arguments for a close link between prediction and intelligence. These include theories about the predictive brain, the relationship between prediction and planning, and the role that prediction plays in artificial intelligence. The following section explores the relationship between an agent's intelligence and its environment and then Section 5 sets out a universal algorithm for measuring predictive intelligence that is based on the interpretation of intelligence developed in the first part of the paper. Experiments to test the practical feasibility of the algorithm are described in Section 6 and the last part discusses potential improvements to the algorithm and some applications.

## 2. Background

### 2.1 Definitions of Intelligence

Many definitions of intelligence have been put forward, including cognitive ability, rational thinking, problem-solving, and goal-directed adaptive behavior (Legg and Hutter 2007a; Neisser et al. 1996;

Yousefian et al. 2016). Intelligence has been linked to accurate prediction (Henaff et al. 2017; Tjøstheim and Stephens 2022), and in the AI community intelligence is often defined as the ability to achieve goals (Legg and Hutter 2007b). People have also suggested that there are multiple *types* of intelligence, such as musical, mathematical, and linguistic intelligence (Gardner 2006). Warwick (2000) generalizes this with his idea that intelligence is a high-dimensional space of abilities.

**2.2 Measurement of Natural Intelligence**

Human intelligence is typically measured with tests that have spatial, numerical and verbal reasoning questions. To calculate IQ, the mean and standard deviation are computed from the test results of a sample of the population. The mean is assigned an IQ of 100 and each standard deviation above and below the mean corresponds to 15 IQ points. So, a person whose test score is the same as the mean has an IQ of 100; a person who scores two standard deviations above the mean has an IQ of 130. The general intelligence factor *g* is calculated by looking for factors that explain correlations between the test results of a sample of the population. Some factors explain correlations between performance on specific intellectual abilities, such as verbal or numerical skills. *g* is a factor that explains correlations between all the different types of test.

IQ and *g* can only be calculated if the population achieves a spread of performance on the tests. If every member of the population gets 0% or 100%, then the standard deviation will be zero and everyone will have the same IQ. In the case of *g*, factor analysis is impossible if all members of the test population achieve the same score. To avoid this problem, the questions used to measure IQ and *g* are carefully selected so that they can be answered with different degrees of success by the target population. Population measures also have the limitation that the numbers that they assign to people's intelligence are not organized on a ratio scale. A person with an IQ of 180 is not twice as intelligent as a person with an IQ of 90.

There has been a substantial amount of work on intelligence testing in animals (Shaw and Schmelz 2017). Some of this has looked at whether animals can solve certain types of problems, such as counting objects or reasoning about hidden causes. Researchers have also measured *g* factor in species, such as mice and chimpanzees (Fernandes et al. 2014; Galsworthy et al. 2014; Galsworthy et al. 2005). However, it is difficult to use population-based measures to compare the intelligence of different species. This would require a set of tests that could be completed to a variable extent by all members of each species.[2] But an effective intelligence test for mice, which could include olfactory discrimination or detour (Locurto et al. 2006), is not appropriate for other species, such as bees or fish. It seems highly unlikely that we will be able to design a single set of tests that could be used to meaningfully compare intelligence across all species.

**2.3 Measurement of Artificial Intelligence**

Human intelligence tests have been used to measure *human-like* intelligence in artificial systems. In early work in this area, Sanghi and Dowe (2003) built an AI system that achieved the same IQ score as an average human. More recently, Campello de Souza et al. (2023) demonstrated that Chat-GPT scores in the 99th percentile on a Brazilian intelligence test. Separate tests for human-like intelligence in AIs have also been developed by Chollet (2019). His ARC questions are used in a $1,000,000 competition, which will award the top prize to an AI that scores 85% (with a $10k compute limit) on the private evaluation set. So far, the highest score that an AI has achieved without a compute limit is 87.5%.[3] The problem with all these tests is that it is not clear whether an AI's performance on the test correlates with its intelligence in other environments, such as the natural and social world. In humans the argument that "IQ tests only measure the ability to take IQ tests" has been addressed by studies that showed correlations between IQ test results and other measures of intelligence, such as success in scientific and professional careers (Naglieri and Bornstein 2003; Robertson et al. 2010). Similar research is required to measure the extent to which an AI's IQ and ARC scores correspond to its intelligence in the natural and social world.

---

[2] IQ and *g* values cannot be compared when they are independently measured in different populations. A tiger with an IQ of 130 is not more intelligent than a human with an IQ of 100.

[3] The highest score within the compute limit is 75.7%. See: https://arcprize.org.

Turing testing can establish whether AIs have the same intelligence as humans. An AI system that was behaviourally indistinguishable from a human could plausibly be attributed the same amount of human-like intelligence. The key limitation of Turing testing is that it can only establish whether a machine has the same level and type of intelligence as a human. AIs with superhuman levels of intelligence will fail Turing tests, as well as AIs that are designed for non-human environments, such as large data sets.

In theory it might be possible to develop separate intelligence tests for AIs that could be used to calculate their IQ and $g$.[4] However, artificial systems are not a clearly defined population (Hernández-Orallo and Dowe 2010), so it will be very difficult to develop a single intelligence test that could be meaningfully administered to all current and future AIs. A second problem with machine intelligence quotients is that most AIs cannot take intelligence tests. Wayve drives cars; AlphaGo plays Go: neither system can take a separate test that would enable their intelligence to be compared on a single scale.

Many methods have been developed to measure the performance of AIs through benchmarks, accuracy metrics, etc. (Hernández-Orallo 2017). However, a measure of performance on a task can only be connected to intelligence if it is clearly specified how the task relates to intelligence. For example, performance on a face recognition data set only measures intelligence if classification of faces is judged to be a task that requires intelligence.

## 2.4 Universal Measures of Intelligence

Universal measures of intelligence have been developed that are, in theory, applicable to any system at all. An influential early proposal was put forward by Legg and Hutter (2007b), who defined intelligence in terms of goal achievement and put forward an algorithm that measures intelligence by summing the rewards that an agent receives across all possible environments, with some adjustment for the complexity of different environments. This measure has some intuitive plausibility, but it is not practically calculable because it sums across all possible actions of the agent in all possible environments. There are also issues with goal achievement and reward in artificial systems. AIs often lack anything corresponding to a human goal and they can arbitrarily change their goals to increase their rewards (a form of wire-heading). The goal achievement interpretation of intelligence also conflates an agent's planning ability with its physical capabilities: weak agents can receive less rewards than equally intelligent strong agents who have a greater ability to manipulate their physical environment (Gamez 2021). A more practical version of Legg and Hutter's (2007b) measure was put forward by Hernández-Orallo and Dowe (2010), which estimates an agent's intelligence from the rewards that it receives from increasingly complex environments. This algorithm is designed to be 'anytime,' so whenever it is halted the result should approximate the system's level of intelligence.

Other universal measures of intelligence have been based on compression. For example, in the Hutter Prize people compete to compress 1GB of Wikipedia data (Hutter 2021), and Hernández-

---

[4] The 'IQ' measures for machines proposed by Bien et al. (2002) and Liu et al. (2017) are based on sets of characteristics that are thought to be linked to intelligence. They are not statistical population-based measures like IQ and $g$.

Orallo's (2000) C-test measures the ability of a system to find the best explanation for sequences of increasing complexity in fixed time. The best explanation is usually a compressed version of the sequences that enables the agent to predict more sequences of the same type.

## 3. Prediction and Intelligence

### 3.1 The Predictive Brain

Suppose that a hunter-gatherer is hungry. They predict that consuming meat will reduce their hunger, and that meat can be found inside a deer. A mental scan of deer habits, the season, water sources, and previous sightings enables them to predict some likely deer locations. Their planning shifts to methods for killing deer, such as rocks, clubs and spears. They decide to use a spear and predict where they can find stones that can be shaped into spear heads, and how pieces of flint will react to blows from a smooth pebble. They predict the shape and size of wood for a shaft that will give the spear strength and the most accurate flight. They predict how different bindings on the spear head (hemp, strips of leather) will resist the impact of the spear on the deer. To hit the deer the hunter predicts the animal's movements and how the trajectory of the spear will be determined by release angle, velocity, windspeed, distance, and so on. These are a tiny part of the predictions that hunter-gatherers make every hour of every day to survive in their environment. Hunter-gatherers that make greater numbers of more accurate relevant predictions will have a survival advantage. This could have contributed to the rapid expansion of the human brain over the last million years.

In the last two decades there has been a surge of interest in the idea that the primary function of the brain is the generation of predictions (Clark 2016; Hohwy 2013). According to these theories, each layer in the mammalian cortex[5] generates predictions about activity in the layer below. Layers compare predictions from higher layers with their own activity and pass prediction errors back up to the layers above. If prediction is a core function of the brain, we would expect there to be a strong correlation between a brain's predictive ability and its intelligence. Detailed theories about the relationships between prediction and natural intelligence have been put forward by Hawkins (2004; 2022) Tjøstheim and Stephens (2022), and Poth et al. (2025).

Predictive brain theories typically treat the brain's predictions as probability distributions. This accommodates situations in which we are certain about something, as well as more common scenarios in which we assign probabilities to different events. People working on the Bayesian brain investigate the extent to which the probability distributions of the brain's representations match the probability distributions of the environment (Knill and Pouget 2004).

### 3.2 Prediction and Planning

Researchers in the AI community often define intelligence as the ability to achieve goals (Legg and Hutter 2007b). Agents that make greater numbers of accurate predictions can make better plans that are more likely to achieve their goals. Consider a person with superhuman predictive ability. They can read the emotions of other people and predict what they are going to say and do next. Their accurate predictions about products and markets would enable them to succeed in business. They would get high scores in exams and intelligence tests (see Section 3.3) and score highly on the societal measures of success that are used to validate IQ tests (Naglieri and Bornstein 2003; Robertson et al. 2010).[6] Now consider a person who cannot predict. They would have little or no understanding of other people and would not be capable of imaginative planning. Basic tool use would be beyond them since they would be unable to predict how objects could help them to solve problems or how materials can be shaped into useful tools. Their inability to predict number sequences or manipulate shapes would give them very low intelligence test scores. At best their behaviour would be limited to reacting to events that happen to them.

---

[5] Predictive brain theories also apply to animals with different brain architectures, such as cephalopods and birds.

[6] This kind of super predictor is dramatized in the film *Knowing* (2009) in which Nicholas Cage can see two minutes into his personal future. Predictive abilities also give protagonists an edge in time travel films, such as *Groundhog Day* (1993) and *About Time* (2013), where people can experience the consequences of their actions and go back in time to try alternatives that are more likely to lead to their goals.

As animals increase in intelligence there is a shift from hard-wired reactions to planned behaviours based on prediction. Snails follow chemical trails and retreat when danger threatens. The world does something to the snail, and it responds in an evolutionarily determined way that, on average, leads to the survival of the species. More sophisticated animals, such as crows and octopi, combine reactive behaviours with actions based on richer predictions about their environment. This enables them to solve more complex problems and build tools. Humans combine their reactive behaviours with planning based on complex predictions on multiple time scales. In all species, tool construction and use depend on accurate predictions about how actions, such as bending a wire, change the shape of the tool and how the finished tool will enable the agent to achieve its goals.

Predictive intelligence only contributes to planning and goal achievement when it is applied in an appropriate area. A person could make excellent predictions about chess while being terrible about planning their way to the supermarket or reading the intentions of a person they want to date. This person's level of predictive intelligence will not correspond to the number of goals they achieve outside chess or to the rewards that they receive from non-chess environments. So, measures of intelligence based on goals and rewards will only correlate with measures based on prediction if the agent's predictions help them to achieve their goals.

### 3.3 Prediction and Intelligence Tests

Many intelligence tests ask the examinee to predict future items in number and shape sequences. For example, in a Raven's matrices test, two sequences of shapes are shown, and the examinee is asked to complete the third. Performance on Raven's matrices is strongly linked to general intelligence (Bartholomew 2004), so people who are better at predicting are likely to achieve higher IQ scores or *g* factors.

### 3.4 Compression

A number of people have suggested that intelligence is linked to a system's ability to compress knowledge, and this has led to universal measures of intelligence based on compression - for example, Hutter (2021). There is a close connection between compression and prediction (Bell et al. 1990), so compression-based theories of intelligence support a link between prediction and intelligence.

### 3.5 Prediction and Artificial Intelligence

Some of the most successful AI systems are based on powerful prediction engines. For example, Deep Blue used a limited search combined with heuristics and knowledge about previous games to predict future game states and choose the best possible moves. Self-driving cars predict how actions, such as pressing the accelerator, affect the car's behaviour under different conditions. They also predict the behaviour of pedestrians, animals, and other drivers, so that they can safely navigate through their environment. Large language models, such as ChatGPT, take a text input and make a prediction about the next sequence of text that is based on what they have learnt.

The main exceptions to this trend are classifiers, such as face recognition algorithms, which are often described as forms of artificial intelligence. Classification could be linked to intelligence through the connection that people make between understanding and intelligence. However, this paper takes the view that classification plays a role in intelligence through its contribution to an agent's ability to predict - it is not a form of intelligence by itself. For example, self-driving cars can predict an object's behaviour more accurately when it has been classified as a pedestrian or car.

**3.6 Spatial and Temporal Prediction**

Predictions are usually interpreted as statements about future events. However, we can also make 'predictions' about events that are simultaneously occurring at inaccessible points in space. At 4pm in London, I 'predict' that people are sleeping in Japan; in my study, I 'predict' that there is a bookcase behind me. In these examples, I am using my intelligence to reach out beyond the spatial boundaries of my senses. We also use our intelligence to discover facts about the past. For example, historians debate the economic and social consequences of the bubonic plague; physicists develop theories about the origins of the universe. This work clearly requires intelligence and is typically called retrodiction.

These cases become straightforward examples of prediction when they are expressed as consequences of agent actions: "*If* I am in Japan at 12:00 GMT, I will observe the majority of people sleeping"; "*If* I turn my head in my study, I will see my bookcase."; "*If* I travel back 14 billion years, I will observe the early stages of the Big Bang." These examples of spatial 'prediction' and retrodiction have been transformed into statements about actions and future observations that an agent could make.

**3.7 The Relationship between Prediction and Intelligence**

The arguments in the previous sections suggest that there is likely to be a close connection between prediction and intelligence. This is formally stated as hypothesis H1:

**H1**. Prediction is the most important component of intelligence.[7]

If H1 is correct, the intelligence of agents will vary with their ability to make accurate predictions:

**H2**. Predictive intelligence varies with the number of accurate predictions that an agent can make.

## 4. Intelligence in Environments and Umwelts

**4.1 The Generality of Intelligence**

Many people believe that intelligence is independent of the environment. For example, John is said to have the same IQ, regardless of whether he is playing chess or relaxing in the cinema. However, real human intelligence does vary with a person's environment. Video gamers have higher levels of intelligence in video game environments than members of Amish communities, who have never used computers. On the other hand, Amish have higher levels of intelligence in agricultural environments. All humans have near zero levels of intelligence in 100-dimensional environments and in environments consisting of petabytes of numerical data. As Chollet (2019) points out, the human brain evolved to help us survive in a hunter-gatherer environment and it has a limited ability to generalize beyond this environment:

> We argue that human cognition follows strictly the same pattern as human physical capabilities: both emerged as evolutionary solutions to specific problems in specific environments (commonly known as "the four Fs"). Both were, importantly, optimized for adaptability, and as a result they turn out to be applicable for a surprisingly greater range of tasks and environments beyond those that guided their evolution (e.g. piano-playing, solving linear algebra problems, or swimming across the Channel) – a remarkable fact that should be of the utmost interest to anyone interested in engineering broad or general-purpose abilities of any kind. Both are multi-dimensional concepts that can be modeled as a hierarchy of broad abilities leading up to a "general" factor at the top. And crucially, both are still ultimately highly specialized (which should be unsurprising given the context of their development): much like human bodies are unfit for the quasi-totality of the universe by volume, human intellect is not adapted for the large majority of conceivable tasks. This includes obvious categories of problems such as those requiring long-term planning beyond a few years, or requiring large working memory (e.g. multiplying 10-digit numbers).

---

[7] H1 is similar to the first thesis put forward by Tjøstheim and Stephens (2022), who claim that natural intelligence "can be approximated as accurate prediction". Their second thesis is not shared by this paper, which rejects the idea of general intelligence (see Section 4.1).

Chollet (2019, pp. 22-3)

IQ tests and the work on *g* have led some people to believe that human intelligence is completely general-purpose. Humans do possess a form of intelligence that enables them to tackle a variety of problems in the cognitive tests used to measure IQ and *g* factor. However, these tests do not prove that humans possess a completely general form of intelligence because both IQ and *g* disappear if the tests are too easy or too challenging (see Section 2.2).

AI researchers often distinguish between *narrow* and *general* artificial intelligence (the latter is often called artificial general intelligence or AGI). Systems that exhibit a single type of intelligence are "narrow" - for example a chess-playing program is a narrow AI because it cannot play draughts or Monopoly. A completely general AI could, in theory, tackle any problem. Human intelligence is often held up as a paradigmatic example of general intelligence, but if human intelligence is "not adapted for the large majority of conceivable tasks", then it is hard to see what a completely general artificial intelligence would look like, and even harder to see how we could build such an intelligence. Without a paradigmatic example of general intelligence, we should consider abandoning the ideal of AGI and compare systems according to the narrowness/generality of their intelligence. Suppose we have ten environments of similar complexity. An intelligence that performs well in eight of these environments is more general (less narrow) than an intelligence that only works in one environment.

If intelligence is, to a greater or lesser extent, specialized, then a measure of an agent's intelligence should be indexed to the set of environments in which it has this level of intelligence. This is formally stated as follows:

> **H3**. An agent's intelligence varies with its environment. If a measure assigns a value to an agent's intelligence, this must be accompanied by the set of environments in which the agent has this amount of intelligence.

Traditional intelligence testing treats intelligence as a property of an agent and assigns a value to this property independently of the agent's environment. This can be expressed as follows, for two agents, $a_1$ and $a_2$:

$$a_1.intelligence = 133$$
$$a_2.intelligence = 963$$

If H3 is correct, the value of an agent's intelligence should be indexed to the environments in which it has this amount of intelligence. Suppose that the intelligence of agent $a_1$ is 133 in environment $E^{43}$ and 58 in environment $E^{109}$. This can be written in the following way:

$$a_1.intelligence^{\{E^{43}\}} = 133$$
$$a_1.intelligence^{\{E^{109}\}} = 58$$

The algorithm described in this paper sums an agent's intelligence across a set of environments (see Section 5). For example, the intelligence of agent $a_2$ could be 3.62 in the *combination* of $E^{215}$ and $E^{1528}$.

This will be represented by placing the set of environments in which the agent has this amount of intelligence inside the curly brackets. For example:

$$a_2.intelligence^{\{E^{215}, E^{1528}\}} = 3.62$$

The relativization of levels of intelligence to sets of environments helps us to understand and appreciate the many different forms that intelligence can take. University professors typically have high levels of intelligence in the academic environments of exams and numerical data. Plumbers can have the same amounts of intelligence as professors in the environments of pipes, valves and water systems that they are familiar with. AIs can have much higher levels of intelligence in high-dimensional numerical datasets than humans. This broader conception of intelligence fits in with previous theories, such as Gardner (2006) and Warwick (2000), and opens us up more fully to the unique forms of intelligence that animals have developed to flourish in their ecological niches.

**4.2 Test and Reference Environments**

Intelligence is typically measured with a special test. The agent is assigned a score on the test, and this is converted into a number that corresponds to the agent's level of intelligence. The agent is then said to have this amount of intelligence in a separate *reference* environment. For example, a person's intelligence is measured using verbal, mathematical, and other questions (the test environment). The test score is converted into an IQ value, which is attributed to the person in the natural and social world (the reference environment).

If human intelligence is completely general, then a person's intelligence will be the same in every environment. A measurement of intelligence in a test environment will be valid in all reference environments. However, the previous section suggested that an agent's level of intelligence changes with their environment (H3). If this is correct, we cannot assume that an agent's level of intelligence in a test environment will be the same as their level of intelligence in a separate reference environment. So, separate intelligence tests can only be used if independent work has shown that the agent's intelligence in the test and reference environments is highly correlated. This is summarized in the following hypothesis:

> **H4**. Valid intelligence tests are supported by measurements of intelligence in the reference environments. If an agent's intelligence is *i* in environment $E^1$, it can only be attributed *i* in $E^2$ if experiments have demonstrated that this type of agent has the same amount of intelligence in $E^1$ and $E^2$.

Several studies have demonstrated correlations between people's intelligence test scores and their academic achievements and success in professional careers (Naglieri and Bornstein 2003; Robertson et al. 2010). So, the level of intelligence that a person achieves in an IQ test *can* be attributed to them in academic and professional environments. However, this does *not* prove that human intelligence is completely general. It only demonstrates that people have the same level of intelligence in IQ tests and in academic and professional environments.

The relationship between test and reference environments is much more problematic for animals and artificial systems. Many cognitive tests have been developed to measure animal intelligence. However, we do not have independent measures of animal intelligence in their reference environments, so it is far from clear how we could prove that intelligence tests in animals measure anything more than their ability to perform the tests. A similar problem exists for AIs, which can be very good at learning regularities in specific environments without developing an ability to generalize beyond those environments. For example, the AI system developed by Sanghi and Dowe (2003) was not able to do anything except pass IQ tests. The problematic relationship between test and reference environments in animals and AIs can be avoided by developing measures of intelligence that can be applied to agents in their reference environments.

### 4.3 The Perceived Environment

Agent terminology is a convenient way to describe the systems whose intelligence is being measured, which can be human, animal or artificial. Agents are often depicted perceiving their environment and carrying out actions in their environment that lead to rewards. This is illustrated in Figure 1a.

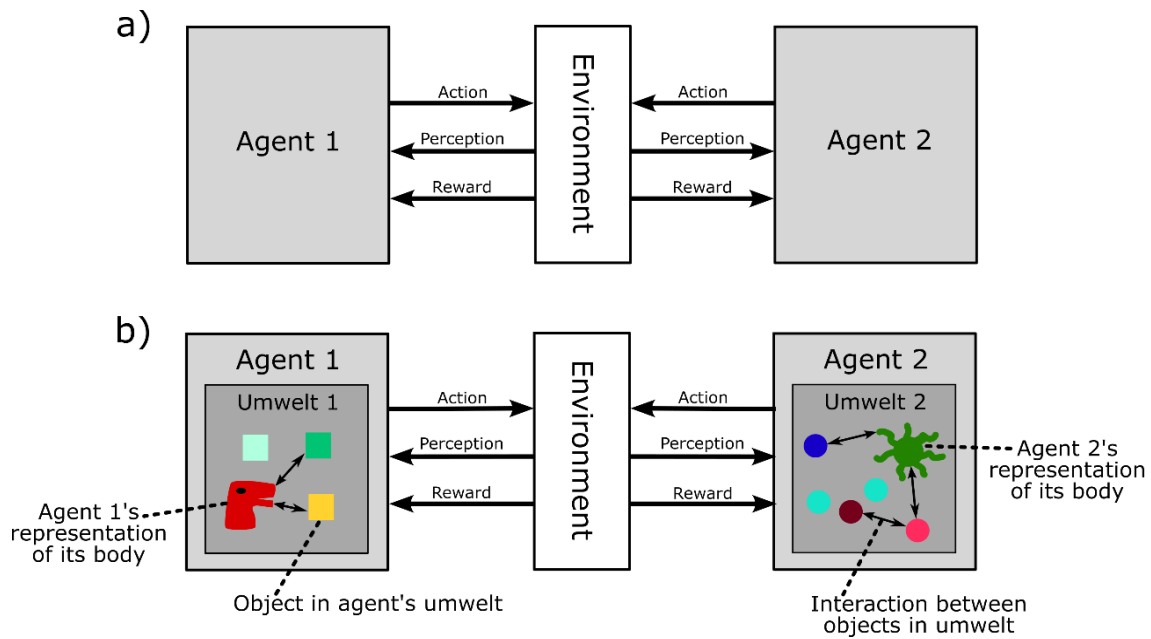

**Figure 1**. Agents and environment. a) Standard agent diagram. The agent perceives its environment and carries out actions in its environment that lead to rewards. b) Sensory data is processed by the agent into a representation of its environment, often known as an umwelt. The agent plans actions in its umwelt, which are executed by the agent's physical or virtual body.

The simple depiction of an agent does not account for the ways in which agents process sensory data into representations of their environments. Suppose a Khepera robot,[8] a dog and a human are in a room. For the Khepera, the states of the environment consist of the distances of objects from its ultrasonic sensors. The dog experiences the room in three-dimensions with limited colour vision and a multidimensional odour map. The human has the best colour perception, one-dimensional odours and a deeper understanding of objects - for example, it can see at a glance that the pattern of colours on the wall is a map.[9] The concept of a perceived environment is nicely captured by the term "umwelt", which is used by ethologists to describe the experiential world of an organism (Uexküll 2010).[10] As an agent interacts with its physical or virtual environment it processes sensory information into representations that appear in its umwelt. The agent plans actions in its umwelt, which are executed by its physical or virtual body. This is illustrated in Figure 1b.

Agents use their intelligence to understand their umwelts and plan actions. For example, the Khepera uses ultrasonic sensor information to map routes around objects – it cannot base its actions on light or chemical information that it cannot sense. The dog knows that a previous canine visitor has come into season, but it cannot use the map to plan a visit to her house. The human understands that they can swat a fly with the paper map, but they have nothing intelligent to say about the neutrino patterns in the room. These observations suggest that an agent's intelligence can only be fully understood in the context of its umwelt. This is summarized in the following hypothesis:[11]

---

[8] There are several versions of the Khepera robot. This example considers a basic version that only has ultrasonic distance sensors.
[9] This example is adapted from Uexküll (2010) and Yong (2022).
[10] Uexküll (2010) uses "umwelten" as the plural of umwelt. This paper uses the more English "umwelts".

> **H5**. Intelligence operates in a perceived environment or umwelt. Two different agents in the same physical environment can be intelligent about different things.

The shift from environments to umwelts has implications for H3 and H4:

> **H3.1**. An agent's intelligence varies with its umwelt. If a measure assigns a value to an agent's intelligence, this must be accompanied by the set of umwelts in which the agent has this amount of intelligence.

> **H4.1**. Valid intelligence tests are supported by measurements of intelligence in the reference umwelts. If an agent's intelligence is $i$ in umwelt $U^1$, it can only be attributed $i$ in $U^2$ if experiments have demonstrated that this type of agent has the same amount of intelligence in $U^1$ and $U^2$.

In this paper $U^1$, $U^2$, … $U^m$ will be used to represent different umwelts. In many cases umwelts are tightly coupled to physical or virtual environments – for example, $U^4$ could be the umwelt that I experience as I move around my physical house; $U^5$ could be the umwelt that I experience as I play Minecraft. Each umwelt has one or more states - for instance, $U^4$ could have states $U^4_1, U^4_2, … U^4_n$. The time at which a state occurs will be indicated by appending the time to the state number after a dash. For example, $U^{46}_{5\text{-}23}$ indicates that state 5 of umwelt $U^{46}$ occurred at time 23.

## 5. A Universal Measure of Predictive Intelligence

This section describes a new universal measure of intelligence based on prediction. The key points are as follows:

- Inspired by internal state-transition algorithms that have been proposed for the measurement of consciousness (Balduzzi and Tononi 2008; Gamez and Aleksander 2011).
- Sums up the accuracy of an agent's predictions (H1, H2).
- Agnostic about the method the agent uses to make predictions.
- Compensates for random guessing.
- Measures intelligence as the agent operates in its ordinary (reference) umwelt (H4.1, H5).
- The value of an agent's intelligence is indexed to a set of umwelts in which it has this amount of intelligence (H3.1).
- Accounts for the complexity of predictions and umwelts.
- Suitable for umwelts with *finite* numbers of states.

### 5.1 Measuring Predictions and Final States

To measure the accuracy of a prediction, we can wait until the time specified in the prediction and then compare the predicted state with the final state. For example, at 15:35 on 23/7/2024 a weather forecast predicts that it will rain in London at 17:00 on 24/7/24. This prediction can be tested by measuring the London weather at 17:00 on 24/7/24.

Predictions are often measured through agents' external behaviour. For example, Alice might say "The weather will be warm tomorrow." This can be tested by waiting until the next day and obtaining another report from Alice about the weather. The measurement of predictions and final states through external behaviour works to a limited extent with agents that speak natural language and with AI systems that are programmed to output their predictions. It is difficult to measure the large numbers of non-verbal predictions that humans and animals make as they interact with their environments. For example, when a monkey reaches for a banana, it is making predictions about the trajectory of its arm, the weight and texture of the banana, the taste of the banana, and so on.

---

[11] Tools and measuring instruments can make hidden properties appear in umwelts. For example, I can say something intelligent about the patterns produced by X-rays when I have made them visible on a photographic plate.

With some agents, the limitations of reporting through external behaviour can be overcome by looking *inside* the agent and identifying two kinds of states:[12]

- Umwelt states.
- States encoding predictions about umwelt states.

Prediction states can be compared with umwelt states at a later point in time to determine whether the predictions came true (see Figure 2). This method can be applied to agents as they interact with their normal reference environments, which avoids the need for a separate test environment.

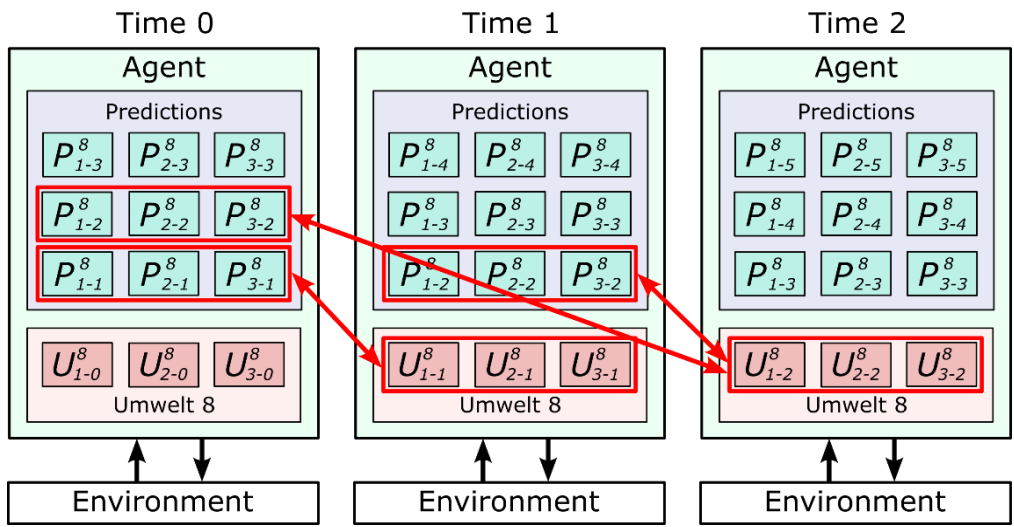

**Figure 2**. Agent's predictions about its umwelt states. The agent is experiencing umwelt 8, which has states $U^8_1, U^8_2$ and $U^8_3$. These could, for example, be face or corner detectors. $P^8_{1-1}$, $P^8_{1-2}$, and $P^8_{1-3}$ are predictions that the agent makes about the values of $U^8_1$ at times 1, 2 and 3. For example, $P^8_{1-1}$ might be the prediction that the face of Jennifer Aniston will appear at time 1 with probability 0.3. As the umwelt changes, predictions are compared with later states to evaluate their accuracy.

State-transition analyses can easily be applied to small- and medium-sized AIs whose internal states are accessible and simple to interpret. With large, opaque and/or complex AI systems, it might be possible to use techniques that have been developed to interpret deep networks (Bau et al. 2020) to identify predictions and internal states. In natural systems there has been a great deal of work on the responses of neurons to features of the environment, such as edges and faces - for example, Quiroga et al. (2005). Theories have also been developed about how predictions are encoded in the brain - for example, Hawkins (2022) – and optogenetics can be used in animals to measure the activity of large numbers of neurons in real time (Lee et al. 2020). This suggests that, in the future, it might be possible to directly measure predictive intelligence from neural activity in animals. Where invasive technologies cannot be applied, predictive intelligence could be estimated from fMRI or EEG, or from biologically accurate large-scale brain simulations. It might also be possible to develop an anytime algorithm that estimates predictive intelligence from external behaviour (see Section 7.1).

---

[12] The measurement of intelligence from internal state transitions is similar to the method that Balduzzi and Tononi (2008) proposed for the measurement of consciousness.

### 5.2 Prediction Accuracy

Prediction accuracy is assessed by comparing predictions with final states. In the most general case, predictions are probability distributions across values that the umwelt state can take and times at which the umwelt state can have these values. For example, Bob might predict that global temperature will rise by 1 ± 0.5 degrees in 10 ± 1 years. For discrete probability distributions, prediction accuracy is measured using Hellinger distance, as shown in Equation 1:

$$H(P,Q) = \frac{1}{\sqrt{2}} \sqrt{\sum_{i=1}^{k} (\sqrt{p_i} - \sqrt{q_i})^2} \qquad (1)$$

Hellinger distance is 0 when there is an exact match between two probability distributions, and 1 when there is a complete mismatch. So 1-*H(P,Q)* gives the *degree of match* between two discrete probability distributions, *P* and *Q*, expressed as a number between 0 and 1.

Some agents make predictions about continuous random variables. Prediction accuracy is measured on this type of system using Student's t-test. If the null hypothesis is accepted, there is a match of 1; otherwise, there is a match of 0. Further work is required to determine the best way of measuring the accuracy of predictions about continuous random variables that are not normally distributed.[13]

### 5.3 Eliminating Random Guesses

Agents can make random guesses about future states. For example, suppose that $U_1^8$ in Figure 2 can have discrete values 6, 7 or 8. In this case, a random guess would be $P(U_1^8=6) = 0.3333$, $P(U_1^8=7) = 0.3333$ and $P(U_1^8=8) = 0.3333$. This guess does not require intelligence because it is generated without knowledge of the environment. However, random guesses can still *partially* match the states that finally occur.

For discrete probability distributions this can be addressed by subtracting the random guesses from the prediction match, as shown in Equation 2.

$$PM(P,U) = |(1 - H(P,U)) - (1 - H(R,U))| \qquad (2)$$

*PM* is the match between a prediction, *P*, and a final umwelt state *U* - for example, the match between $P_{1-2}^8$ and $U_{1-2}^8$ in Figure 2. *R* is an equal distribution across possible sensor values. Random guessing can be ignored for continuous random variables since the agent has a close to zero chance of guessing a particular value.

### 5.4 Exploration of Environments and Umwelts

To get the total prediction match in an umwelt, the algorithm puts the agent through every possible state transition and compares each of the agent's predictions with the states that finally occur. In most cases, it is difficult to manipulate the umwelt directly, so the umwelt is altered by changing the agent's physical or virtual environment. Agents can be manually controlled, or they can be allowed to explore freely, and the observer can wait until every possible umwelt state transition has been encountered. The summation of prediction matches is given in Equation 3:

$$PM^u = \sum_{s=1}^{y} \sum_{t=1}^{z} PM(P_{s-t}^u, U_{s-t}^u) \qquad (3)$$

---

[13] Gamez (2023) added error bounds to the final umwelt state and calculated the prediction match from the intersection between the prediction and the error bound box.

The prediction matches are summed for all states of the umwelt (*s=1…s=y*), and for all the time indexed predictions about each state (*t=1…t=z*). This method of measuring predictive intelligence has the important limitation that it only works with umwelts that have a *finite* number of possible states. In the future, an anytime version of this algorithm could be developed that estimates an agent's intelligence from a sample of its umwelt states (see Section 7.1).

### 5.5 Complexity of Predictions

Agents can generate large numbers of trivial predictions about simple umwelts. For example, suppose an umwelt has a single state which always has a value of 1. It would not take much intelligence to generate an infinite number of accurate predictions about this umwelt's future states. To address this issue, the sum of the prediction matches is multiplied by the Kolmogorov complexity of the predictions, as shown in Equation 3.1:

$$PM^u = \frac{K(p)}{L(p)} \sum_{s=1}^{y} \sum_{t=1}^{z} PM(P_{s-t}^u, U_{s-t}^u) \qquad (3.1)$$

*PM$^u$* is the total prediction match for the umwelt *U$^u$*. *p* is a string that describes all the predictions in *U$^u$*. *K(p)* is the Kolmogorov complexity of *p*, and *L(p)* is the length of *p*. The predictions are summed for all umwelt states (*s=1…s=y*), and for all time indexed predictions (*t=1…t=z*). Kolmogorov complexity cannot be directly calculated, so it is usually approximated with compression algorithms.[14]

### 5.6 Intelligence in Multiple Umwelts

Predictive intelligence is relative to a *set* of umwelts (H3.1), so *PM$^u$* is summed up for all the agent's umwelts. This raises the problem that the concept of a distinct umwelt is not clearly defined.[15] Some umwelts are very different; others have trivial variations between them. If the *PM$^u$* values from different umwelts were simply added together, an agent would double its intelligence across two umwelts that are almost identical. To address this issue, the sum of *PM$^u$* across umwelts *U$^1$*, *U$^2$*, … *U$^x$* is multiplied by the Kolmogorov complexity of the combined umwelts divided by the sum of the Kolmogorov complexity of the umwelts considered independently, as shown in Equation 4:

$$PM^{\{U^1…U^x\}} = \frac{K(U^1 + U^2 + \cdots + U^x)}{K(U^1) + K(U^2) + \cdots + K(U^x)} \sum_{u=1}^{x} PM^u \qquad (4)$$

*K* is Kolmogorov complexity and $PM^{\{U^1…U^x\}}$ is the total prediction match for umwelts *U$^1$*… *U$^x$* after the differences between the umwelts have been accounted for.

If two umwelts are very similar, the complexity of the combined umwelts will be approximately the same as their individual complexity, and the first half of Equation 4 will approximate 1/2. On the other hand, if two umwelts are very different, then the length of the shortest program describing the umwelts together will be approximately equal to the sum of the lengths of the shortest programs describing the umwelts individually. In this case the first half of Equation 4 will be close to 1. In practice, Kolmogorov complexity is approximated using compression algorithms, which need to be carefully chosen to take the nature of the umwelts into account.

---

[14] One potential issue with Equation 3.1 is that the precision of predictions can affect their complexity. For example, if the final value of an umwelt state is 0.25, a prediction of 0.2560121342344 has greater complexity than a more accurate prediction of 0.251. This can be addressed by rounding the numbers prior to the complexity calculation.

[15] The concept of a distinct environment is also not clearly defined.

### 5.7 Comparing Simple and Complex Agents

The log is taken of the total prediction match $PM^{\{U^1...U^x\}}$ to make it easier to compare complex agents, such as humans, with trivial AI systems on a single scale. This also reduces the effect of decisions about how the algorithm is implemented. The log of a very small number is a large negative number, which makes no sense for intelligence, so predictive intelligence is set to zero if $PM^{\{U^1...U^x\}}$ is less than or equal to 1. This is summarized in Equation 5:

$$ᚹ_{1.1}^{\{U_1...U_x\}} = \begin{cases} log_2(PM^{\{U^1...U^x\}}) & if\ PM^{\{U^1...U^x\}} > 1 \\ 0 & otherwise \end{cases} \quad (5)$$

The letter that has been selected to represent this measure of intelligence is ᚹ, which is the Old Norse letter (rune) that corresponds to our modern 'p' sound. ᚹ is pronounced "peorth", "perth" or "pertho". The ᚹ rune is associated with the dice cup, chance, secrets, destiny and the future, which is appropriate for a measure that is based on a system's ability to make predictions. The superscript is the set of umwelts in which the agent has this level of predictive intelligence. The subscript is the version of the algorithm.[16]

## 6. Experiments

Preliminary experiments were carried out to establish whether $ᚹ_{1.1}$ could be a practical way of measuring intelligence in real AI systems. The systems selected for these experiments were an embodied agent that learnt to predict the consequences of its actions in virtual maze environments, and an agent that used a neural network to make predictions about time-series data.

### 6.1 Maze Agent

The maze agent is intended as a simple example of an embodied agent that interacts with a real or virtual environment. The agent has four sensors: S1-S3 provide information about the square to the left, front and right of the agent; S4 has information about the contents of the square that the agent is occupying. The maze states can be wall (W), empty (E) or reward (R). In these experiments reward is just an extra feature – the agent does not 'want' the reward or gain anything from being in a square with the reward. The mazes are shown in Figure 3.

---

[16] Version 1.0 of the algorithm is described in Gamez (2023).

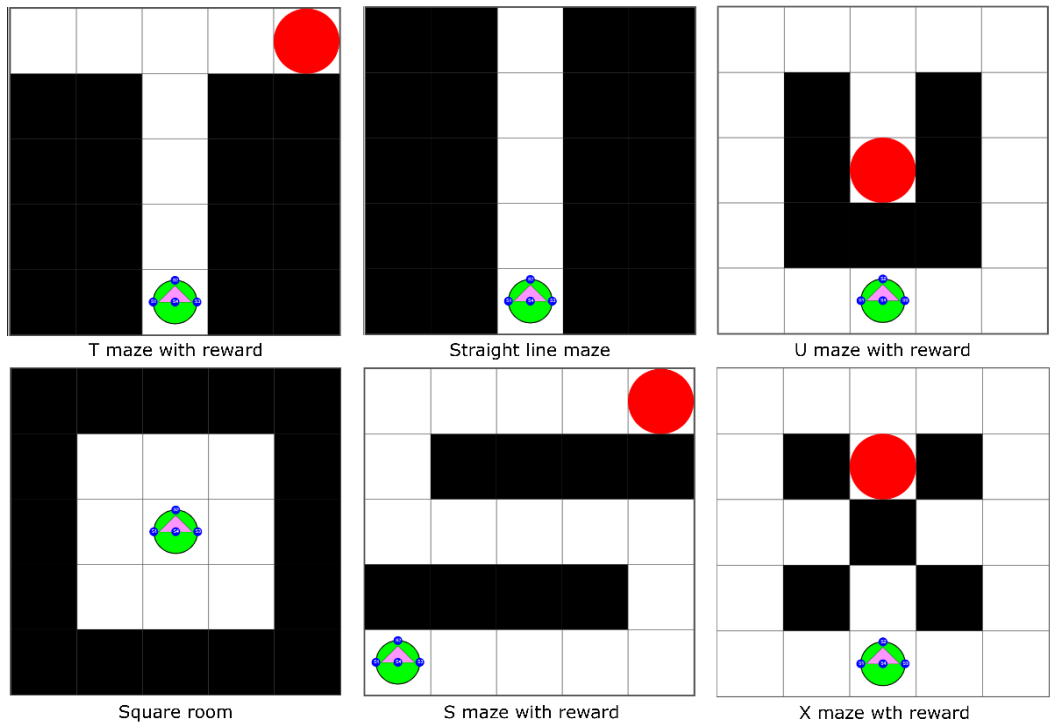

**Figure 3**. Mazes used in the experiments. The green circle is the agent; the red dot is the reward.

The agent can move forward one square and rotate to point in four different directions (up, down, left, right). The user can control the agent manually or the agent can automatically explore all possible actions at each location. Table 1 gives some examples of how the agent's sensor states change in response to its actions.

| | | Current State | | | Action | Next State | | | |
|---|---|---|---|---|---|---|---|---|---|
| | S1 | S2 | S3 | S4 | | S1 | S2 | S3 | S4 |
| T maze with reward | W | E | W | E | Move | W | E | W | E |
| | W | E | W | E | Left | E | W | E | E |
| | E | W | E | E | Down | W | E | W | E |
| | W | E | W | E | Move | W | W | W | E |
| X maze | E | E | E | E | Move | W | W | W | E |
| | W | W | W | E | Left | E | W | W | E |
| | E | W | W | E | Move | E | W | W | E |

**Table 1**. Illustration of sensor-action combinations for two maze environments. In the T maze the agent moves forward, points left, points down and then moves forward, so that it ends up at the starting position pointing downwards. In the X maze, the agent moves forward, points left and tries to move forward again. It cannot do this because of the wall, so it remains in its current position and the sensors remain unchanged. W=wall; E=empty square.

As the agent explores a maze it records the state transitions that it has encountered (see examples in Table 1). This data is used to make simple statistical predictions about the sensory consequences of its actions. For example, in the T maze with reward, suppose the agent moves forward four times from its starting position shown in Figure 3. The first three moves result in WEWE states. The final move into the crossbar of the T will result in a EWEE state. So, an agent that has moved from the bottom to the top of this maze will make the prediction shown in Table 2 when it next tries to move forward from a WEWE state:

| Sensor | P(W) | P(E) | P(R) |
|--------|------|------|------|
| S1 | 0.75 | 0.25 | 0 |
| S2 | 0.25 | 0.75 | 0 |
| S3 | 0.75 | 0.25 | 0 |
| S4 | 0 | 1 | 0 |

**Table 2**. Prediction about the next sensor state that will occur if the agent tries to move forward after four moves up the T maze.

The agent does not build maps or mental models of the maze that correspond to the images shown in Figure 3. Instead, the umwelt corresponding to a maze can be loosely described as the set of sensor-action combinations that occurs in the maze. For the purposes of testing the algorithm, string descriptions of the mazes were used as proxies for the agent's umwelts, and the complexity of the umwelts and predictions were approximated with a LZUTF8 compression algorithm. Random guessing was implemented by assigning equal probability to the three possible states, so for each sensor $P$(W)=0.33, $P$(E)=0.33 and $P$(R)=0.33.

To evaluate the agent's intelligence, learning was switched off and the agent was automatically made to execute every possible action at each location. $𝕂_{1.1}$ was then calculated for the mazes separately and in combination. The maximum intelligence was evaluated by assuming that the agent made the best possible statistical predictions about the sensory consequences of each action.

**6.2 Time-series Agent**

A second set of experiments was carried out with an agent that used a neural network to make predictions about time-series data. The data sets used in these experiments are shown in Figure 4.

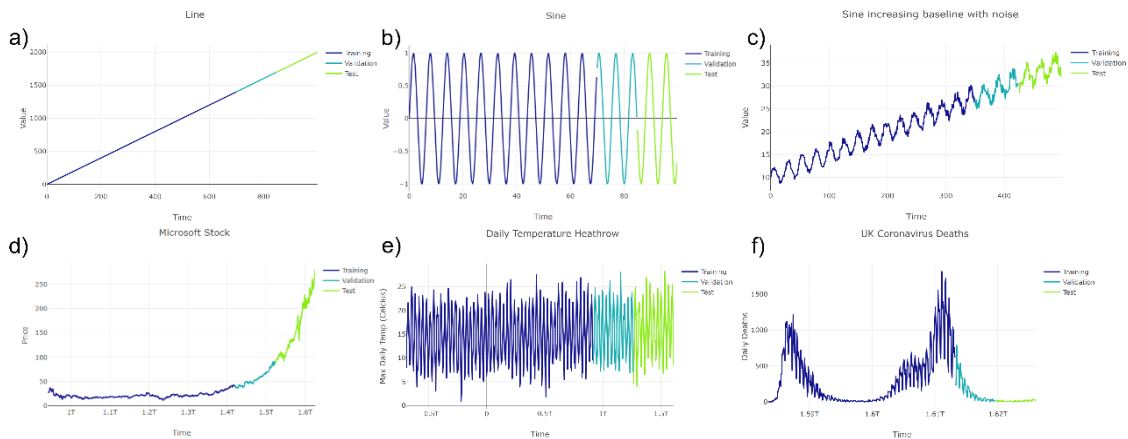

**Figure 4**. Data used with the time-series agent. a) Line; b) Sine; c) Sine with increasing baseline and noise; d) Microsoft stock price; e) Maximum daily temperature at Heathrow Airport, London, UK; f) Daily deaths from coronavirus in the UK. Each data set was divided into training, validation and test sections.

The data sets were split into short overlapping sequences using a sliding window. These were fed into LSTM cells, which were densely connected to an output layer that was trained to predict the next number in the sequence after the sliding window. To generate a spread of predictions, separate networks were trained on each data set and their predictions were combined into a normal distribution around the mean. The prediction match was then calculated using Student's t-test (see Section 5.2). The hyperparameters used in the experiments are given in Table 3.

| Hyperparameter | Value |
|---|---|
| Time window | 3 |
| Number of models/networks | 5 |
| LSTM units per model | 20 |
| Number of epochs | 10 |
| Batch size | 10 |
| p-value for prediction match | 0.05 |

**Table 3**. Hyperparameters used in time series experiments. These can be set by the user through the web interface (see Figure 5).

The networks in this agent have internal states that might correspond to some kind of umwelt that is distinct from the input data fed into the network. Techniques for visualizing the representations learnt by deep neural networks could be used to identify these umwelts – for example, see Bau et al. (2020) - although it is unlikely that they would reveal much in such simple networks. For the purposes of demonstrating the algorithm, this step was omitted, and the compressibility of the data sets (calculated using LZUTF8) was used as an approximation to the compressibility of the networks' umwelts.

In machine learning experiments the data is typically divided into train, validation and test sets (a typical split is 70% training, 15% validation and 15% test). The system is trained on the train set, fine-tuned on the validation set and the final performance is evaluated on the test set, which the system has not encountered before. This methodology is used to develop systems that can apply what they have learnt to novel data. The experiments to evaluate $\mathcal{I}_{1.1}$ needed to measure the complete intelligence of the agent in each data environment. So, the networks were trained with the train sets, fine-tuned on the validation sets and then $\mathcal{I}_{1.1}$ was calculated for the entire data set.[17] The maximum possible intelligence was estimated by assuming a perfect match for each prediction.

To get a rough idea of the performance of the algorithm, the intelligence calculation times were measured for the maze agent and time-series agent over 20 runs for different numbers of predictions. These tests were conducted on an AMD Ryzen 7 3800X CPU, which ran the intelligence calculations without GPU acceleration. The timings for the maze agent included the time to generate the predictions. The duration of the time-series intelligence calculations did not include training and querying the neural network, which was supported by GPU.

The experiments are implemented on a website (see Figure 5), which is available at: www.davidgamez.eu/pi. The source code can be downloaded from this website.

---

[17] A network that has overfitted on a data set can be highly intelligent in this data set. However, it is likely to have less intelligence in novel data sets than a network that has generalized effectively from the data it has been exposed to.

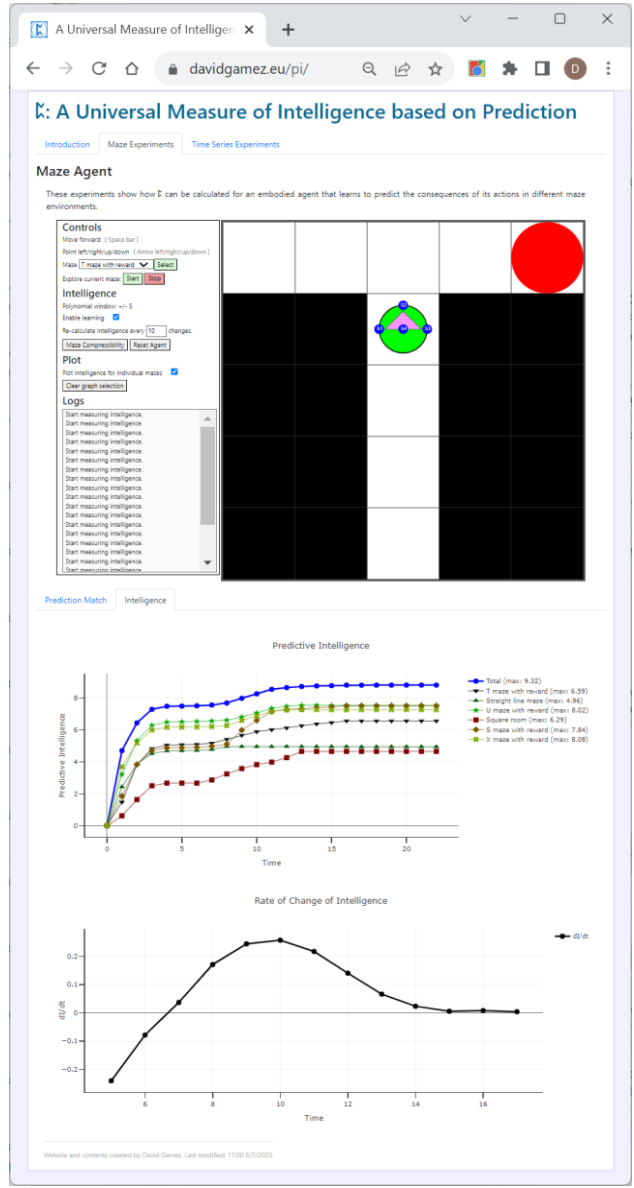

**Figure 5.** Website with code and experiments.

### 6.3 Results

Illustrative runs of the maze agent and time-series agent are plotted in Figure 6. These show the changing predictive intelligence of the agents after learning is switched on and they interact with different environments. Tables 4 and 5 give information about the size of the environments, their compressibility and maximum values of $ꓘ_{1.1}$. The maximum values of $ꓘ_{1.1}$ that were achieved by both agents were close to the theoretical maximums, despite the simple implementations of intelligence in the agents. This is partly due to the logarithmic component of the algorithm and partly because intelligence was calculated across the entire environment. The similarity between the final values of $ꓘ_{1.1}$ for the maze and time-series agents is also partially due to the logarithm, and roughly similar numbers of predictions are possible in the two environments. Although the agents have similar levels of intelligence, the umwelts in which they have these levels of intelligence are very different: the maze agent has zero intelligence in the time series environments; the time-series agent has zero intelligence in the maze environments.

David Gamez

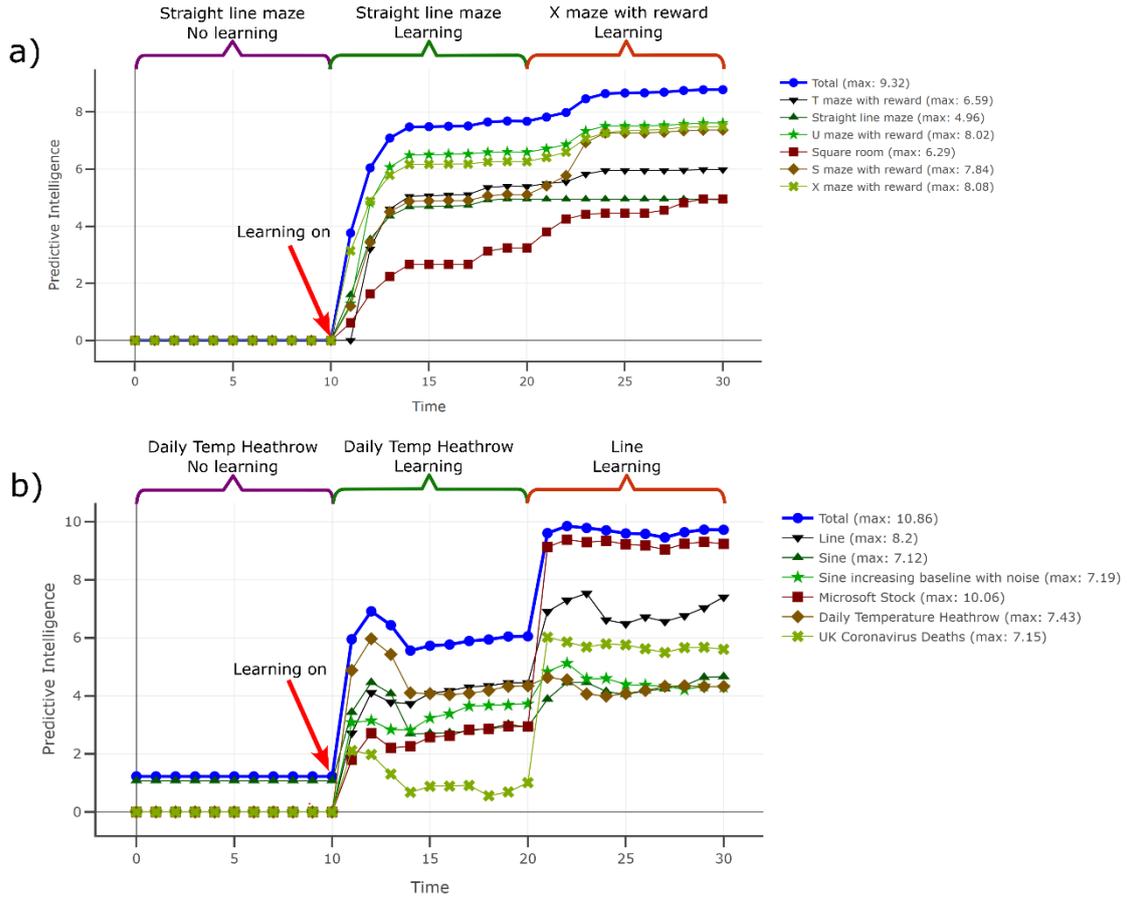

**Figure 6**. Illustrative plots of the agents' intelligence as they interact with different environments. a) $\mathcal{K}_{1.1}$ of maze agent. When learning is switched on the agent learns about the statistical properties of the straight line maze, leading to an increase in its intelligence. $\mathcal{K}_{1.1}$ increases again when it is placed in a new maze. b) $\mathcal{K}_{1.1}$ of time-series agent. When learning is switched on the time-series agent trains on the Heathrow temperature data set and $\mathcal{K}_{1.1}$ increases for this data set and similar environments. $\mathcal{K}_{1.1}$ increases again when the agent is exposed to the line data set. This also increases its intelligence in the Microsoft stock dataset, which has a large linear component.

| Maze | Number of Actions | Compressibility | Max $\mathcal{K}_{1.1}$ |
|---|---|---|---|
| T-maze with reward | 144 | 0.48 | 6.59 |
| Straight line maze | 80 | 0.28 | 4.96 |
| U maze with reward | 288 | 0.64 | 8.02 |
| Square room | 144 | 0.4 | 6.29 |
| S maze with reward | 272 | 0.6 | 7.84 |
| X maze with reward | 320 | 0.6 | 8.08 |
| **All mazes** | **1248** | **0.37** | **9.32** |

**Table 4**. Maximum possible values of $\mathcal{K}_{1.1}$ for the maze agent in each maze considered individually and across all mazes. This table also shows the number of possible actions (move and change direction in different locations) in each maze and their compressibility, which was calculated individually and collectively using the LZUTF8 algorithm.

| Data Set | Number of Items | Compressibility | Max ꓘ$_{1.1}$ |
|---|---|---|---|
| Line | 1000 | 0.86 | 8.21 |
| Sine | 500 | 0.76 | 7.12 |
| Sine increasing baseline with noise | 500 | 0.80 | 7.19 |
| Microsoft stock price | 5457 | 0.63 | 10.06 |
| Daily temperature Heathrow | 876 | 0.57 | 7.43 |
| Coronavirus deaths UK (CD) | 494 | 0.79 | 7.16 |
| **All data sets** | **8827** | **0.65** | **10.87** |

**Table 5**. Maximum possible values of ꓘ$_{1.1}$ for the time-series agent on each data set considered individually and across all datasets. This table also shows the number of items in each data set and their compressibility, calculated individually and collectively using the LZUTF8 algorithm.

The performance results shown in Figure 7 suggest that the time to calculate ꓘ$_{1.1}$ is likely to scale linearly with the number of predictions, which is consistent with the equations in Section 5. The divergence between timings for the maze and time-series agents is largely due to implementation differences. In the current setup with no code optimization or GPU acceleration it should take around 1.25 hours to calculate ꓘ$_{1.1}$ on a system that makes a billion predictions.

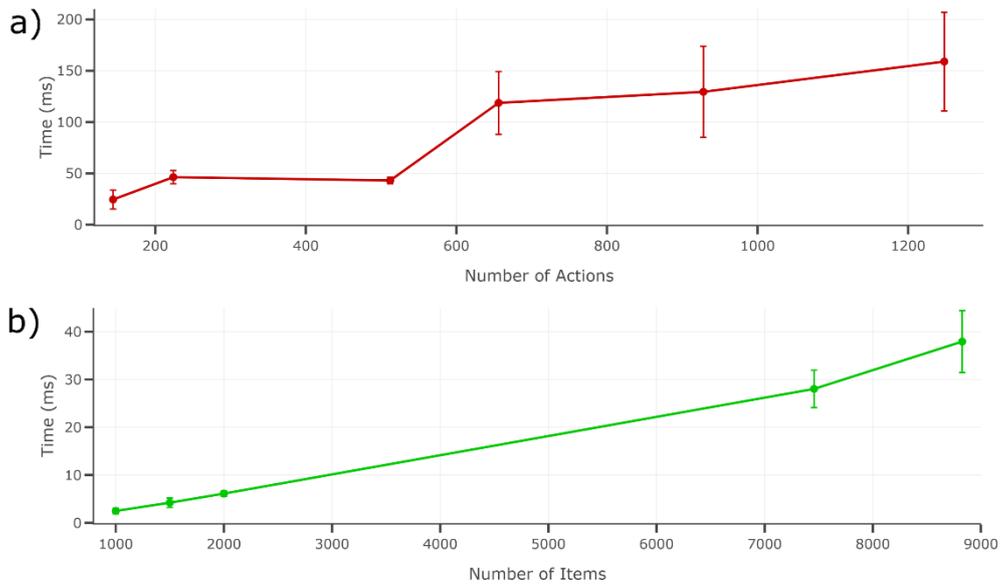

**Figure 7**. Timing of ꓘ$_{1.1}$ calculations for different numbers of predictions. Each data point is the average over 20 runs. Error bars are +/-1 standard deviation. a) Timings of intelligence calculations for maze agent. b) Timings of intelligence calculations for time-series agent.

## 7. Discussion

### 7.1 Improvements to the Algorithm

This paper has presented a linear algorithm that can completely measure the predictive intelligence of small- to medium-sized AI systems. GPU acceleration should reduce the computation time by a few orders of magnitude, but it will not be possible to completely measure the intelligence of large systems that interact with complex environments, such as Deep Blue ($10^{40}$ chessboard states) or AlphaGo ($10^{170}$ Go board states). To measure the intelligence of these agents it will be necessary to develop an *anytime* algorithm that can estimate ꓘ$_{1.1}$ from a subset of the umwelt states - similar to the algorithm

David Gamez

developed by Hernández-Orallo and Dowe (2010). One option would be to use a Monte Carlo method that measures the prediction match at random points in each umwelt and scales this up to estimate $\Bbbk_{1.1}$. The accuracy of an anytime algorithm could be evaluated by comparing its results to $\Bbbk_{1.1}$ on smaller systems.

$\Bbbk_{1.1}$ measures intelligence from state transitions inside the agent. The advantage of this approach is that the intelligence of AIs can be assessed without the need for separate tests. However, with current technology, it is difficult to apply this approach to humans and to non-transparent biological systems (see Section 5.1). A first step towards measuring the predictive intelligence of biological systems would be to apply an anytime version of $\Bbbk_{1.1}$ to embodied neural simulations of animal brains – for example, the rat brain developed by Aldarondo et al. (2024) or the C-elegans model created by the OpenWorm project (Szigeti et al. 2014). Data from neural simulations could be combined with reports, observations of external behaviour, and specially designed tests to generate rough estimates of the predictive intelligence of different species.

Further work is required to adapt $\Bbbk_{1.1}$ to work with analogue systems that have infinite numbers of possible states. The intelligence of these systems could be approximated by converting the continuous values into discrete values with an adjustable level of precision, or the algorithm could be rewritten to integrate intelligence over the umwelt. $\Bbbk_{1.1}$ would also have to be adapted to handle discrete umwelts with infinite numbers of states. For example, agents can be immersed in procedurally generated game environments that extend forever with minor variations. One potential solution would be to divide the umwelt into finite sections and measure the changing intelligence of the agent as more parts are added. With procedurally generated environments, the joint complexity term of Equation 4 should reduce the prediction match contributions from later sections, and $\Bbbk_{1.1}$ should converge to a finite value. Further work is required to determine if this approach could work with natural environments and with environments that contain significant amounts of randomness.

## 7.2 Applications

Statistical measures of intelligence, such as IQ and *g*, are based on questions that are designed to elicit a spread of responses from a target population. This approach cannot be used to compare the intelligence of different populations, such as humans, animals and artificial systems. Previous universal measures of intelligence are either impractical or depend on separate test environments. As far as I am aware, $\Bbbk_{1.1}$ is the first practical universal measure of intelligence that can be applied directly to an agent as it interacts with its normal environment. This could be the starting point for a new comparative science of intelligence that ranks humans, animals and AIs on a single ratio scale.

Concerns are often raised about artificial intelligence exceeding human intelligence, leading to some kind of hostile or benevolent takeover by AIs (Bostrom 2016; Russell 2019). However, if humans do not have general intelligence, then it is unclear how AGIs could emerge or be built. Dangerous AIs *without* general intelligence could emerge if the following conditions are met:

1) The AI is in an environment in which it can do bad things to humans. For example, it is connected to the Internet, where it could hack into critical infrastructure and cause damage that kills people – for example, opening a dam.
2) The AI's umwelt contains rich representations of aspects of the environment that are relevant to doing bad things to humans. AIs with rich representations of cats are unlikely to threaten humans. AIs with rich representations of networking, security protocols and physical infrastructure are potential threats.

3) The AI is more intelligent than humans in the relevant umwelts.
4) The AI can act in its environment.
5) The AI has a reason (something like a goal or reward) to act in its environment or could act in its environment through a system error.

When the first two conditions are met, ꓘ$_{1.1}$ could be used to measure the AI's intelligence, and steps could be taken to limit its intelligence or its ability to cause harm.

ꓘ$_{1.1}$ could be used to develop more advanced cognitive systems. For example, it could serve as feedback to a genetic algorithm or reinforcement learning system, which could use this signal to autonomously develop higher levels of intelligence. This intelligence could be combined with an emotion system that associates positive and negative valence with umwelt states (Damasio 1994). The agent could then use its predictive intelligence to plan actions that lead to umwelt states with positive valence.

According to Popper (2002), the strength of a scientific theory corresponds to its ability to make falsifiable predictions: strong scientific theories make large numbers of falsifiable predictions; weak scientific theories make few falsifiable predictions. A measure of prediction accuracy and quality like ꓘ$_{1.1}$ could potentially be used to compare the strength of scientific theories. It could also help us to discover better scientific theories - for example, by judging competing theories or evolving new theories, as part of work on computational scientific discovery (Sozou et al. 2017).

## Conclusions

The first part of this paper set out the following interpretation of intelligence:

1. *Prediction is the most important component of intelligence (H1, H2).* Several arguments were put forward for a strong link between prediction and intelligence. A predictive interpretation of intelligence overlaps with goal-based definitions of intelligence and aligns with recent theories about prediction and the brain.
2. *Intelligence operates in a perceived environment or umwelt (H5).* There is a great deal of variability of agents' sensors and the ways in which they process sensory information into representations of their environments. Agents apply their intelligence to their perceived environments.
3. *An agent's level of intelligence varies with its perceived environment (H3.1).* All the intelligences that we know are, to a greater or lesser extent, specialized for one or more umwelts. So, an agent's level of intelligence should be indexed to the set of umwelts in which it has this amount of intelligence.
4. *Valid intelligence tests are supported by measures of intelligence in the reference umwelt (H4.1).* Most measures of intelligence are based on separate tests that are different from the agent's usual (reference) umwelt. If intelligence is not completely general, then it cannot be assumed that the agent will have the same level of intelligence in the test and reference umwelts. So, intelligence tests must be supported by experiments that demonstrate a correlation between the agent's intelligence in the test and reference umwelts. This work has been done for some human umwelts. We do not have independent measures of animal or AI intelligence in their reference umwelts.

This interpretation of intelligence led to the algorithm for measuring intelligence that was described in the second part of the paper. As the agent explores its environment, the states of its umwelt change, and it makes predictions about future states of its umwelt. The algorithm sums up the accuracy of these predictions at every possible state of each umwelt, and then the prediction match is combined for a set of umwelts. Kolmogorov complexity is used to compensate for trivial predictions and for similarities between umwelts. Finally, the logarithm is taken to make it easier to compare simple and complex systems.

Two sets of experiments were carried out to determine the practical feasibility of the algorithm. The results demonstrated that $к_{1.1}$ could be measured in an embodied agent and in an agent that interacts with a data environment. $к_{1.1}$ scales linearly with time and is calculable on medium sized AI systems. To analyse larger systems, it will be necessary to develop an anytime version of the algorithm that can estimate $к_{1.1}$ from small amounts of data and progressively refine the estimate as the agent is exposed to more of its environment. It will also be necessary to develop methods for estimating $к_{1.1}$ from external behaviour and in umwelts that do not have finite numbers of states.

In the future, $к_{1.1}$ could contribute to a shift in the study of intelligence, from population-based statistics to a universal science that compares the intelligence of humans, animals and AIs on a single ratio scale. $к_{1.1}$ could also help us to understand and mitigate AI threats, contribute to computational scientific discovery, and help us to develop more intelligent cognitive systems.

ꟼ: A Universal Measure of Predictive Intelligence